\documentclass[letterpaper, 10 pt, conference]{ieeeconf}
\usepackage{cite} 
\usepackage{multirow}
\usepackage[utf8]{inputenc}
\usepackage[misc]{ifsym}
\usepackage{graphicx}
\usepackage{tikz}
\usepackage{pifont}
\usepackage[misc]{ifsym}
\usepackage{tikz}
\usepackage{pdfpages}
\usepackage{graphicx}
\usepackage{amsmath}
\usepackage[utf8]{inputenc}
\usetikzlibrary{arrows,shapes,positioning,shadows,trees,calc,decorations.markings}

\usepackage{amssymb}
\usepackage{hyperref}
\usepackage{booktabs}
\usepackage{algorithm}
\usepackage{algpseudocode}
\usepackage{balance}

\usepackage{color, colortbl}
\definecolor{LightCyan}{rgb}{0.88,1,1}

\hypersetup{
colorlinks=true,
linkcolor=black
}

\makeatletter
\newcommand{\printfnsymbol}[1]{%
  \textsuperscript{\@fnsymbol{#1}}%
}
\makeatother

\IEEEoverridecommandlockouts                

\title{\LARGE \bf
ETSM: Automating Dissection Trajectory Suggestion and Confidence Map-Based Safety Margin Prediction for Robot-assisted Endoscopic Submucosal Dissection
}

\author{Mengya Xu$^{1,2}$\printfnsymbol{1}\thanks{\printfnsymbol{1} Equal contribution}, Wenjin Mo$^{3}$\printfnsymbol{1}, Guankun Wang$^{1,2}$\printfnsymbol{1}, Huxin Gao$^{1,2}$, An Wang$^{1,2}$, Long Bai$^{1,2}$, \\
Chaoyang Lyu$^{4}$, Xiaoxiao Yang$^{4}$, Zhen Li$^{4}$, and Hongliang Ren$^\dagger$$^{1,2}$
\thanks{$^\dagger$ This work was supported by Hong Kong RGC CRF C4026-21GF, GRF (14203323, 14216022, \& 14211420), NSFC/RGC Joint Research Scheme N\_CUHK420/22, Shenzhen-HK-Macau Technology Research Programme (Type C) STIC Grant 202108233000303. (Corresponding to: H. Ren, hlren@ieee.org.)}
\thanks{$^{1}$ Department of Electronic Engineering, The Chinese University of Hong Kong, Hong Kong SAR, China.}
\thanks{$^{2}$ Shenzhen Research Institute, The Chinese University of Hong Kong, Shenzhen, China.}
\thanks{$^{3}$ Department of Computer Science and Engineering, Sun Yat-sen University, Guangzhou, China.}
\thanks{$^{4}$ Department of Gastroenterology, Qilu Hospital of Shandong University, Jinan, China.}
}

\begin{document}
\maketitle
\thispagestyle{empty}
\pagestyle{empty}

\begin{abstract}
Robot-assisted Endoscopic Submucosal Dissection (ESD) improves the surgical procedure by providing a more comprehensive view through advanced robotic instruments and bimanual operation, thereby enhancing dissection efficiency and accuracy. Accurate prediction of dissection trajectories is crucial for better decision-making, reducing intraoperative errors, and improving surgical training. Nevertheless, predicting these trajectories is challenging due to variable tumor margins and dynamic visual conditions. To address this issue, we create the ESD Trajectory and Confidence Map-based Safety Margin (ETSM) dataset with $1849$ short clips, focusing on submucosal dissection with a dual-arm robotic system. We also introduce a framework that combines optimal dissection trajectory prediction with a confidence map-based safety margin, providing a more secure and intelligent decision-making tool to minimize surgical risks for ESD procedures. Additionally, we propose the Regression-based Confidence Map Prediction Network (RCMNet), which utilizes a regression approach to predict confidence maps for dissection areas, thereby delineating various levels of safety margins. We evaluate our RCMNet using three distinct experimental setups: in-domain evaluation, robustness assessment, and out-of-domain evaluation. Experimental results show that our approach excels in the confidence map-based safety margin prediction task, achieving a mean absolute error (MAE) of only $3.18$. To the best of our knowledge, this is the first study to apply a regression approach for visual guidance concerning delineating varying safety levels of dissection areas. Our approach bridges gaps in current research by improving prediction accuracy and enhancing the safety of the dissection process, showing great clinical significance in practice. The dataset and code are available at \href{https://github.com/FrankMOWJ/RCMNet}{https://github.com/FrankMOWJ/RCMNet}.
\end{abstract}



\section{Introduction}
\label{sec:intro}

Endoscopic Submucosal Dissection (ESD) is a surgical technique used to manage early-stage gastrointestinal cancers~\cite{chiu2012endoscopic,zhang2020symmetric}. This procedure involves a series of intricate dissection actions that necessitate a high level of expertise to identify the most effective dissection path. Providing insightful recommendations for dissection trajectories holds significant promise for providing decision-making assistance and enhancing surgical skills training~\cite{kim2011factors}. In addition, the dissection path suggestion technology can also be integrated into an image-guided surgical robot prototype that can automatically assist the surgeon and perform a single or a series of surgical steps in the robotic endoscopic surgical procedure. Nevertheless, predicting the optimal dissection trajectory for future stages is notably challenging. The complexity arises from various factors, such as the safety margins surrounding the tissue. Moreover, dynamic surgical scenes can further hinder the accurate evaluation of the dissection trajectory.


Despite its clinical significance, predicting dissection trajectories during Endoscopic Submucosal Dissection (ESD) has been relatively understudied in research~\cite{guo2020novel,qin2020davincinet,wang2022real,li2023imitation}. A notable exception is the recent development of implicit diffusion policy imitation learning (iDiff-IL)~\cite{li2023imitation}, which leverages video demonstrations of expert procedures to generate dissection trajectory predictions.
Similarly, the long short-term memory (LSTM) model~\cite{wang2022real} has been introduced for real-time prediction of laparoscopic instrument tip trajectories. Another relevant approach is inspired by pedestrian trajectory prediction, which views motion indeterminacy diffusion (MID)~\cite{gu2022stochastic} as a reverse process. This method systematically reduces uncertainty across walkable areas to pinpoint the desired trajectory. 

\begin{figure*}[!hbpt]
\centering
\includegraphics[width=0.7\linewidth]{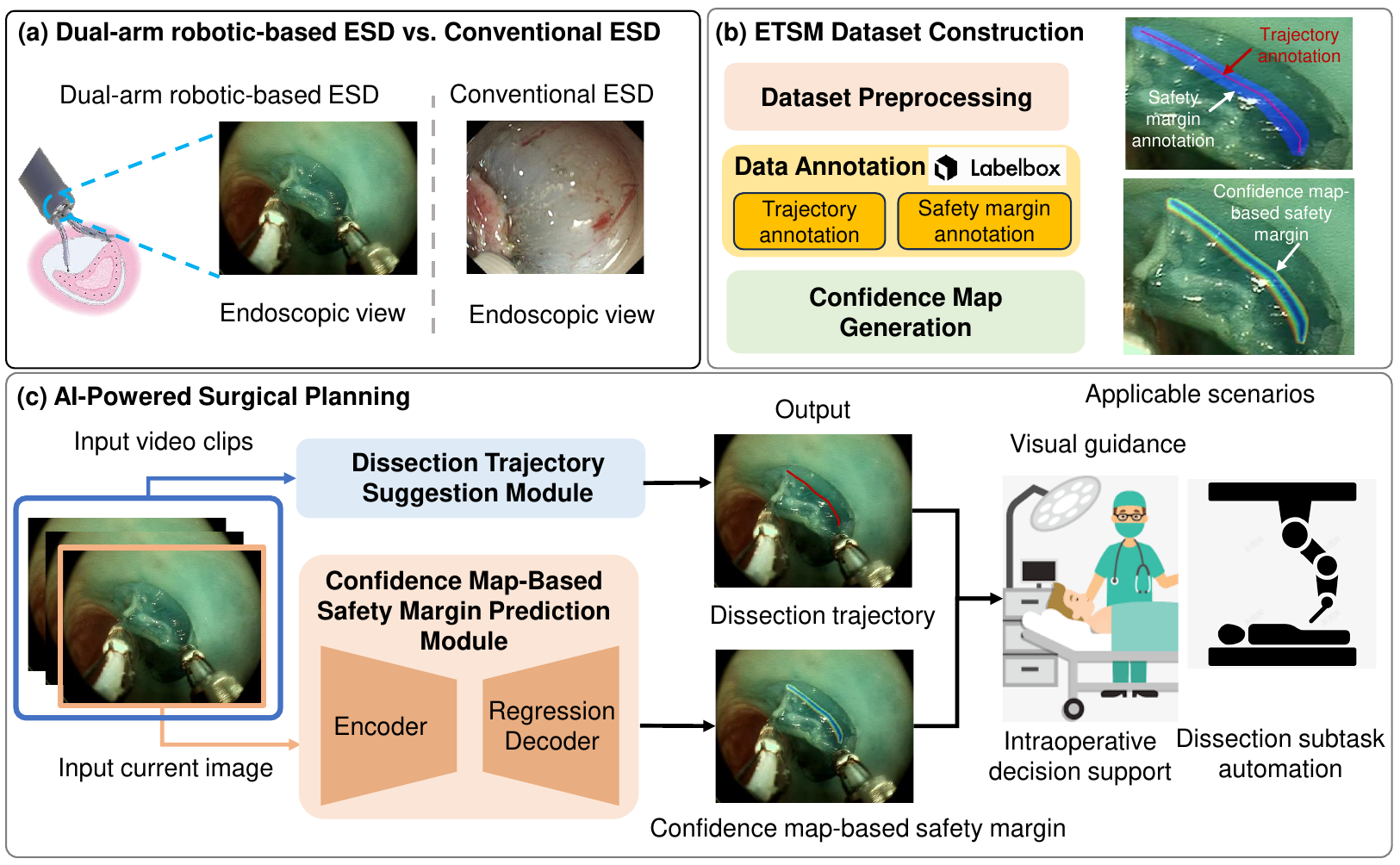}
\caption{Overview of our workflow. (a) \textbf{The dual-arm robotic-based ESD vs. Conventional ESD:} The dissection trajectory in the limited endoscopic view during conventional ESD is partial. Dissection trajectory for robot-assisted ESD is more complete. (b) \textbf{ETSM Dataset Construction:} The dataset preprocessing involves video downsampling, frame extraction, and removal of black margins. Data annotation includes marking dissection trajectories with a series of 2D coordinates and annotating safety margins. The ground truth confidence map for the dissection area is generated based on the optimal dissection trajectory and safety margin annotations. (c) \textbf{AI-powered surgical planning system:} includes two functional modules: the dissection trajectory suggestion module and the confidence map-based safety margin prediction module. The output provides intraoperative decision support through visual guidance and may also facilitate future dissection subtasks automation.}
\label{fig:overview}
\vspace{-0.6cm}
\end{figure*}

However, most works do not consider the safety margin around the dissect trajectory when predicting the optimal dissect trajectory. This leads to uncertainty regarding the extent of deviation between the predicted dissection trajectory and the actual optimal dissection trajectory, as well as the surgical safety margin boundaries. The most satisfactory submucosal dissection requires the removal of the lesion or mucosal layer with the complete submucosal layer while maintaining the lesion or mucosal layer and muscular layer intact. The misoperation of the electric knife may cause muscular injuries or even perforations, which increases the surgical risks and affects the postoperative recovery~\cite{gao2024transendoscopic}. Besides, the mucosal layer is also likely to be damaged by the knife despite the protection of cushioned submucosa, which will reduce the accuracy of histopathological examination for the lesion. Consequently, such a safety margin in real-time instruction for endoscopists plays an important role in increasing ESD safety. Therefore, this work proposes to predict the optimal dissection trajectory and confidence map-based safety margin as the more intelligent and secure decision-making aid.

In this work, we advance surgical guidance and automation by developing a novel dissection trajectory suggestion framework and associated tools. This work aims to enhance surgical precision and efficiency, ultimately improving patient outcomes. Our contributions can be summarized as follows:
\begin{itemize}
\item We introduce a comprehensive framework integrating dissection trajectory suggestion with confidence map-based safety margin prediction. This dual-component approach not only provides surgeons with critical visual guidance during procedures but also paves the way for the potential automation of dissection tasks. 

\item We create the ETSM dataset (\textbf{E}SD \textbf{T}rajectory and Confidence Map-based \textbf{S}afety \textbf{M}argin). This dataset, derived from ESD videos captured by the DREAMS (Dual-arm Robotic Endoscopic Assistant for Minimally Invasive Surgery) system~\cite{gao2024transendoscopic}, includes detailed annotations of dissection trajectories and safety margins. 

\item We first propose to apply a regression approach for visual guidance in delineating varying safety levels in dissection areas and design the \textbf{R}egression based \textbf{C}onfidence \textbf{M}ap prediction \textbf{Net}work (RCMNet), which contains a Transformer-based image encoder and an All-MLP regression decoder. 

\item We propose an angular-difference-based algorithm to generate the confidence map-based safety margin. Such confidence distribution provides critical guidance for the surgeon during the operation.
\end{itemize}

\section{Methodology}
\subsection{ETSM Dataset Construction} 
Our \textbf{ETSM} dataset, which stands for \textbf{E}SD \textbf{T}rajectory and confidence map-based \textbf{S}afety \textbf{M}argin, is constructed from the following steps: (1) extracting images from the ESD video clips of the submucosal dissection task, (2) labeling the dissection trajectory and the map-based safety margin, and (3) generating confidence distributions for dissection regions based on trajectory and margin data (see Fig.~\ref{fig:overview}(b)).

\textbf{Image Collection from ESD Videos.} The first stage of our proposed pipeline is the acquisition of surgical videos using the DREAMS system. The animal study was approved by the Institutional Ethics Committee (Approval No. DWLL-2021-021). We utilized a custom robotic platform~\cite{gao2024transendoscopic, yang2024novel} to record 21 robotic ESD procedures on ex-vivo porcine models. This work focuses on the submucosal dissection task due to its extensive soft tissue interaction and procedural complexity. Video sequences related to this task are selectively clipped and downsampled to 1 FPS, forming the dissection dataset. The operator interface is cropped to produce a final image resolution of 1310×1010. Due to the continuous movement of the surgical instruments and the camera, the dissection region is frequently occluded or misaligned with the camera view, rendering predictions of the dissection area in such cases impractical. To address this, we manually filter keyframes from the whole dataset, ensuring a good and unobstructed view of the dissection region.

\textbf{Annotation Comparisons between Robot-assisted and Conventional ESD.}
Conventional ESD uses a transparent cap to retract lesions, often resulting in an obstructed view of the submucosal layer and incomplete dissection trajectories. In contrast, robot-assisted ESD enhances surgical precision by eliminating the need for endoscope manipulation through dexterous robotic instruments and allowing separate control of retraction and dissection. This approach provides a clearer view of the submucosal layer, leading to more complete dissection trajectories and increased efficiency and accuracy in procedures (see Fig.~\ref{fig:overview}(a)).

\textbf{Dissection Trajectory and Map-Based Safety Margin Annotation.}
In clinical practice, endoscopists must select an optimal dissection trajectory in real-time and precisely control the robotic electric knife to follow this trajectory for submucosal dissection. The ideal trajectory is defined by three key criteria: \textit{efficacy, safety, and efficiency}. \textit{Efficacy:} The trajectory should ensure complete removal of the submucosal layer, with the knife precisely following the boundary between the submucosal and muscular layers. To account for potential manipulation errors, the trajectory is adjusted to offset by 0.4 mm, the diameter of the knife tip. This offset is estimated by skilled endoscopists during data labeling due to the lack of real size representation in endoscopic views. \textit{Safety:} The trajectory must avoid areas with potential safety issues, such as poor illumination or occlusion. By adhering to these criteria, an optimal dissection trajectory is determined. Additionally, to enhance labeling reliability, endoscopists annotate safety margins. \textit{Efficiency:} Effective tissue retraction is crucial for efficient dissection. Thus, trajectories in areas with insufficient traction are excluded to ensure quicker dissection. 

    


\begin{figure}[t]
    \centering
    \includegraphics[width=0.8\linewidth, trim=0 30 0 0]{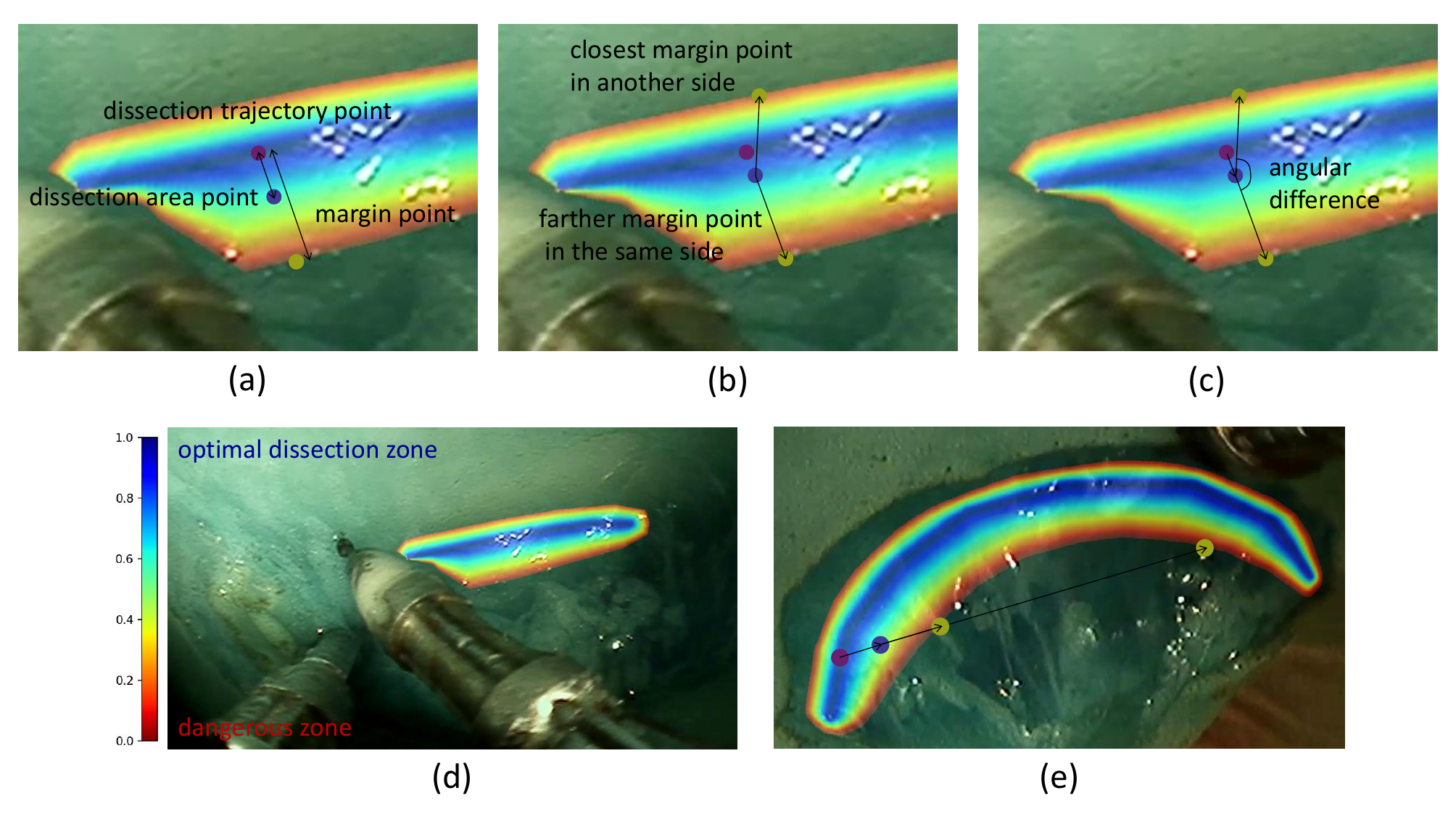}
    \caption{Case illustrations of confidence generation. (a) The point in the dissection area depends on the distance of its location from the optimal dissection trajectory and edge. (b) The search is easily misled by the other side of the edge when searching for the edge point through the smallest Euclidean distance. (c) Searching for edge points by angular difference can avoid edge mislead while allowing the confidence to transit in a fixed direction. (d) Full view of this case's dissection area confidence. (e) For a curved dissection area, it is necessary to add a threshold on the Euclidean distance between the area point and the edge point.}
    \label{Fig_CG}
\vspace{-0.6cm}
\end{figure}

\textbf{Generation of Confidence Distributions for Dissection Area.}
Based on the annotations of the dissection trajectory (DT) and safety margin, we can infer both the optimal dissection position and the dissection area (DA). The confidence distribution, transitioning from the optimal dissection position to the margin, provides critical guidance for the surgeon during the operation. To generate the confidence distribution, we assign a confidence level of 1 to the DT position and a confidence level of 0 to both the safety margin and its exterior. Consequently, the confidence values within the DA smoothly decrease from 1 to 0, thereby delineating varying safety levels of dissection areas. Specifically, as shown in Fig.~\ref{Fig_CG}(a), areas closer to 1 (shown in darker blue) indicate higher safety levels for dissection, while areas closer to 0 (depicted in alert red) represent lower safety levels. The transition zones between them reflect varying safety levels for dissection. For a seamless transition, the confidence at any DA position is determined by the ratio of its distances to the nearest point of the DT and the safety margin (see Fig.~\ref{Fig_CG}(a)). However, a naive distance-based approach for nearest point search can lead to inaccuracies, as it may select the closest margin point on the opposite side of the area point relative to the DT (see Fig.~\ref{Fig_CG}(b)), resulting in an erroneous confidence estimation. To address this, we propose an angular-difference-based algorithm to locate the optimal margin point (see Fig.~\ref{Fig_CG}(c)). For any point in the DA, the nearest trajectory point is first identified as a calibration point. The ideal confidence distribution should exhibit a directional transition from the DT to the safety margin. Therefore, we cast a ray from the calibration point through the area point, which serves as the reference direction. By casting rays from the area point to all points on the safety margin and comparing their angular differences with the reference direction, we can select the optimal margin point with the minimum direction angular difference. This margin point is then used for the confidence calculation. Our angular-based algorithm ensures that the selected margin point $(x_e, y_e)$ and the trajectory points $(x_t, y_t)$ are nearly collinear with the area point $(x_a, y_a)$, and the confidence is calculated as follows:
\begin{equation}
\begin{aligned}
C = \frac{\sqrt{(x_e - x_a)^2 + (y_e - y_a)^2}}{\sqrt{(x_t - x_a)^2 + (y_t - y_a)^2}},
\end{aligned}
\end{equation}
where $C$ is the confidence in any area point. In some instances, the DA exhibits a curved shape, potentially leading to cases where the edge point with the smallest angular difference is located at a considerable distance from the area point, and the confidence exceeds 1 (see Fig.~\ref{Fig_CG} (d)). To address this issue, a distance threshold between the margin point and the area point is introduced.

\subsection{AI-Powered Surgical Planning System}
The surgical planning system comprises two primary functional modules. The dissection trajectory suggestion module assists in planning optimal dissection trajectories, while the confidence map-based safety margin prediction module evaluates and visualizes safety margins (see Fig.~\ref{fig:overview}(c)).


\subsubsection{\textbf{Dissection Trajectory Suggestion}}
The goal of the dissection trajectory prediction is to generate a reliable future dissection trajectory based on the previous observation. The input is a video clip consisting of frames $s = {f}_{t-L+1}, {f}_{t-L+2},..., f_{t}$,  $f_{t} \in \mathbb{R} ^{H\times W \times C}$, and the output is a series of 2D coordinates \( C = \{c_{t+1}, c_{t+2}, \ldots, c_{t+N}\} \), with \( c_t \in \mathbb{R}^2 \), representing future dissection trajectories.

We employ behavior cloning (BC)~\cite{codevilla2019exploring}, a fully supervised method utilizing a CNN-MLP network. Additionally, we have also chosen motion indeterminacy diffusion (MID)~\cite{gu2022stochastic}, a diffusion-based trajectory prediction technique and implicit diffusion policy imitation learning (iDiff-IL)~\cite{li2023imitation}, representing the current state-of-the-art.


\subsubsection{\textbf{Confidence Map-Based Safety Margin Prediction}}
\begin{figure}
    \centering
    \includegraphics[width=0.85\linewidth, trim=0 30 0 0]{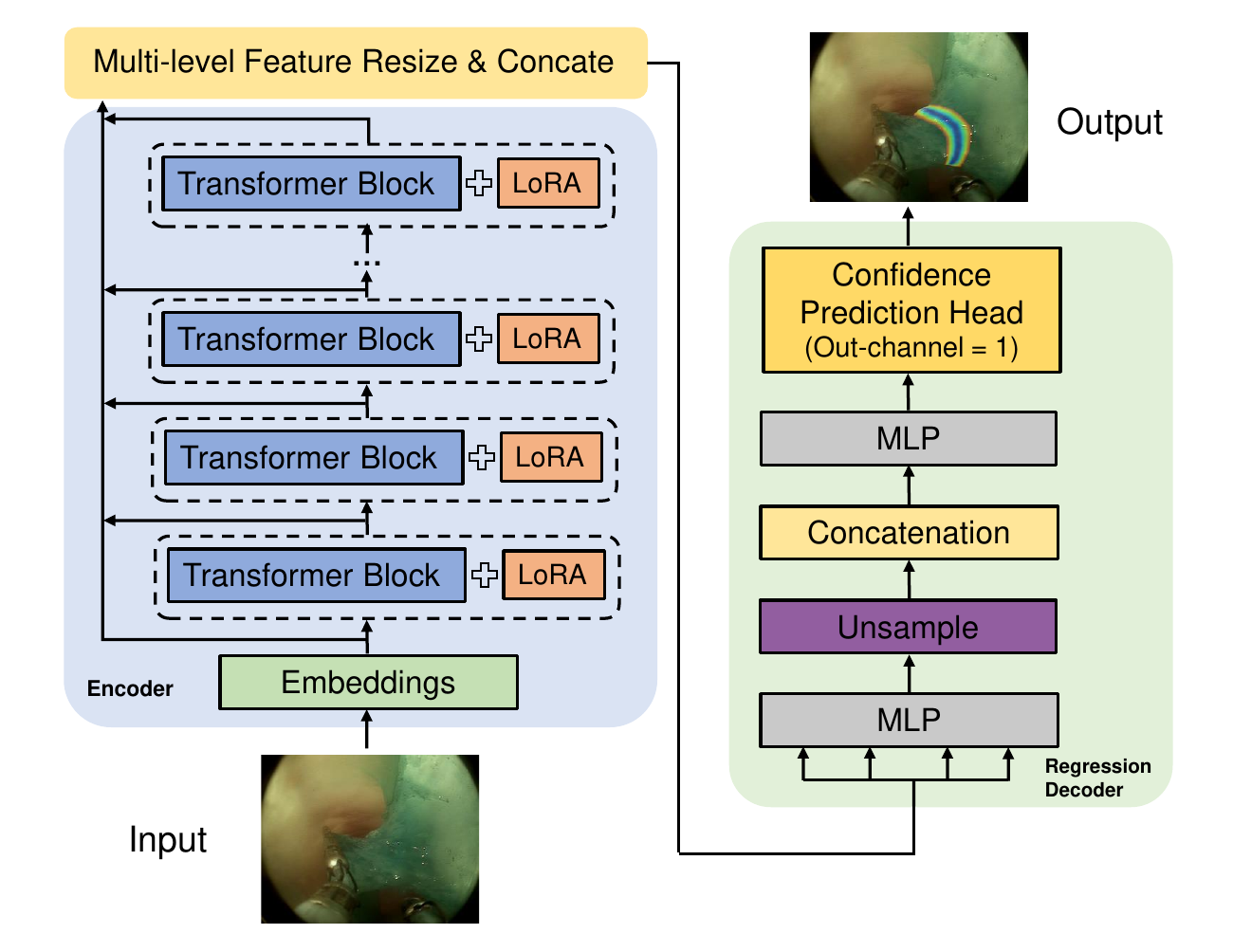}
    \caption{Overview of our RCMNet. The image encoder, leveraging a pre-trained DINOv2, extracts multi-scale feature representations from intermediate transformer layers. Features from each layer are concatenated along the channel dimension and upsampled by a factor of 4. These multi-scale features are then fed into a regression decoder based on an ALL-MLP network. The decoder first aligns channel dimensions via an MLP, upsamples the features back to the input resolution, and fuses them through another MLP. Finally, an MLP confidence prediction head is used to generate a 1-channel confidence map.
 }
    \label{Fig_Arch}
\vspace{-0.6cm}
\end{figure}

As illustrated in Fig.~\ref{Fig_Arch}, our model can be conceptually split into an encoder module and a decoder module. The encoder is responsible for processing the input image and the decoder reads from the encoder output to generate the confidence map.

\paragraph{\textbf{Image Encoder}}
The objective of the image encoder is to extract latent representations of raw images. DINOv2~\cite{oquab2023dinov2} is a pre-trained image encoder that can generate both high-level and pixel-level vision features. We adopt the pre-trained DINOv2 as our image encoder, applying LoRA layers to each Transformer block for efficient fine-tuning and capturing task-specific learnable information. For an image $x \in \mathbb{R} ^{H\times W \times C}$, it is first to be separated into non-overlapping patches $x_t^0 \in \mathbb R ^ {N \times D}$,$1 \leq t \leq N$, where $N = \frac{HW}{p^2}$, $p$ is the size of patch and $D$ is the dimension of each patch. Also, a class token $x_0^0 \in \mathbb R ^{1 \times D}$ is added to the image embeddings and forms the input embeddings $x^0 = \left\{ x_0^0, ..., x_N^0\right\}$. Then, the embeddings are encoded by $K$ transformer blocks to extract feature representations $x^k$, the intermediate feature map of the $k^{th}$ transformer block. We utilized the Vit-Base model from DINOv2 with 12 transformer blocks, patch size $p$ of 14 and a feature dimension $D$ of 784.
Intermediant outputs from  $k = \left\{ 3, 6, 9, 12\right\}$ transformer blocks are extracted as image representations. The class token and image tokens from the same layer are concatenated along the channel dimension and then upsampled by a factor of 4.

\paragraph{\textbf{Regression Decoder}}

To decode the image features to the confidence map, we explore three types of decoders.  
\begin{itemize}
    \item \textbf{Single Convolution Layer.} We use a single convolution layer to transform the feature map to a 1-channel output.

    \item \textbf{Fully Convolutional Network(FCN).} The Fully Convolutional Network (FCN) demonstrates excellent performance in generating dense predictions by leveraging its capability to capture spatial hierarchies and contextual information. Following the approach described in \cite{DBLP:journals/corr/LongSD14}, the input features are processed through $N$ convolutional layers. The output from these layers is then concatenated with the original input features, and a subsequent convolutional layer is employed to fuse the combined features. Finally, a 1-channel output convolutional layer acts as the confidence map decoder, transforming the features into the final prediction.

    \item \textbf{All-MLP Network.} Inspired by Segformer\cite{xie2021segformer}, RCMNet  utilizes an All-MLP network as a decoder, which has been demonstrated to be a simple, lightweight, yet powerful and efficient decoder for use with Transformer architectures.  There are five stages in the decoder. First, the multi-level features from the DINOv2 encoder are processed through an MLP layer to unify the number of channels. Second, the features are upsampled back to the size of the inputs of the image encoder. Third, the features are concatenated and then fused by another MLP layer. Finally, an additional MLP layer, serving as a confidence map decoder, takes the fused features to generate the 1-channel confidence maps.
    
\end{itemize}

\paragraph{\textbf{Loss Function}}
To punish the model when predicting non-zero confidence out of safety margin, we propose weight Mean Square Error (MSE). The weight MSE of one predicted confidence map $C \in \mathbb{R}^{H \times W \times 1}$ can be formulated as
\begin{equation}
         \mathrm{MSE}_{\text {weighted }}=\frac{1}{H \times W} \sum_{i=1}^{H} \sum_{j=1}^{W} w_{ij}\left(C_{ij}^{\mathrm{gt}}-C_{ij}\right)^{2},
\end{equation}
where $C^{\mathrm{gt}}$ is the corresponding ground truth. $w_{ij}$ is the loss weight of pixel $C_{ij}$ which is set to 10 when $C_{ij}^{\mathrm{gt}} = 0$,
indicating that the pixel $C_{ij}$ is outside the safety margin.
Otherwise, we set $w_{ij}=1$, giving equal importance to all pixels.

\section{Experiments}
\subsection{Our ETSM Dataset}

\begin{figure*}[h]
    \centering    
    \includegraphics[width=0.7\linewidth]{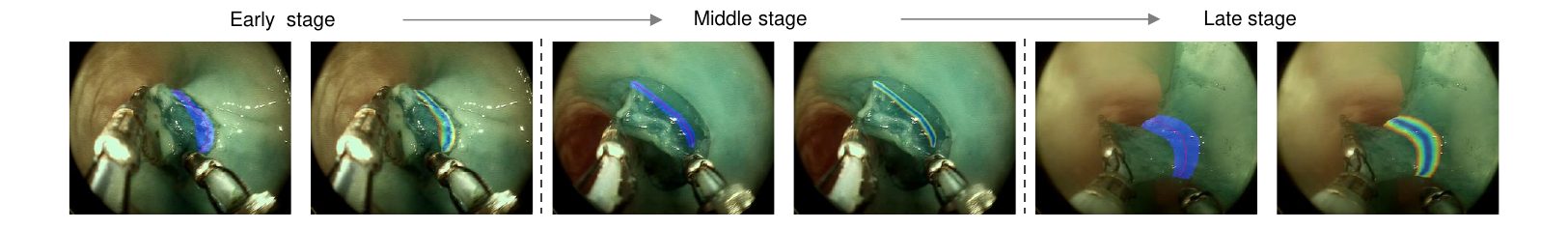}
    \caption{Our ETSM dataset visualization. Frames are selected from videos of submucosal dissection tasks performed using a dual-arm robotic system.}
    \label{fig:dataset}
\vspace{-0.2cm}
\end{figure*}

\begin{table*}[htbp]
  \centering
  \caption{Confidence map-based safety margin prediction results. For models with poor task performance, we do not consider them in the robustness evaluation.}
  \scalebox{0.7}{
    \begin{tabular}{c|c|cc|ccc|cccc|cc|cccc|c}
    \toprule
    \multirow{3}[2]{*}{Model} & \multirow{3}[2]{*}{Resolution} & \multicolumn{2}{c|}{In-domain evaluation} & \multicolumn{14}{c}{Robustness evaluation using MAE} \\
\cline{5-18}          &       & \multirow{2}[2]{*}{MAE $\downarrow$} & \multirow{2}[2]{*}{MSE $\downarrow$} & \multicolumn{3}{c|}{Noise} & \multicolumn{4}{c|}{Blur}     & \multicolumn{2}{c|}{Weather} & \multicolumn{4}{c|}{Digital}  & \multicolumn{1}{c}{Mean} \\
          &       &       &       & Gauss. & Shot  & Impulse & Speckle & Defocus & Motion & \multicolumn{1}{p{4.2em}|}{Zoom} &  Fog  & Bright & Constrast & Elastic & Pixel & JPEG  & MAE $\downarrow$\\
    \hline
    MAN   & \multirow{5}[2]{*}{224*224}      & 3.74  & 657.22 & -      & -      & -      & -      & -      & -      & -      & -      & -      & -      & -      & -      & -      & - \\
    SASNet &       & 4.04  & 424.7 & 5.16  & 5.16  & 5.17  & 4.77  & 4.06  & 4.34  & 4.33  & 5.32  & 4.06  & 3.85  & 4.54  & 4.02  & 4.19  & 4.54 \\
   Our RCMNet-v1 &       & 5.16  & 414.80 & 5.23  & 5.34  & 5.30  & 5.17  & 5.25  & 5.59  & 5.63  & 6.04  & 5.43  & 5.58  & 5.98  & 5.05  & 5.30  & 5.45 \\
    Our RCMNet-v2 &       & 4.57  & 398.57 & 5.06  & 5.23  & 4.96  & 4.79  & 4.95  & 5.34  & 4.96  & 4.51  & 4.26  & 4.27  & 5.39  & 4.35  & 4.74  & 4.83 \\
    Our RCMNet &       & \textbf{3.46} & \textbf{378.95} & 4.78  & 4.60  & 4.86  & 4.14  & 4.10  & 4.45  & 4.25  & 4.13  & 3.73  & 3.89  & 4.71  & 3.76  & 4.10  & 4.27 \\
    \hline
    MAN   & \multirow{5}[2]{*}{532*532}      & 3.68  & 643.40 & -      & -      & -      & -      & -      & -      & -      & -      & -      & -      & -      & -      & -      & - \\
    SASNet &       & 4.19  & 437.4 & 4.72  & 4.53  & 5.31  & 4.51  & 3.93  & 4.21  & 4.17  & 5.23  & 4.18  & 3.77  & 4.52  & 4.04  & 4.26  & 4.41 \\
    Our RCMNet-v1 &       & 3.87  & 356.80 & 6.08  & 5.92  & 6.45  & 5.08  & 5.68  & 5.25  & 5.33  & 5.56  & 4.08  & 4.98  & 5.34  & 4.97  & 5.17  & 5.38 \\
    Our RCMNet-v2 &       & 3.46  & 352.64 & 4.02  & 3.99  & 3.87  & 3.89  & 4.09  & 4.04  & 4.14  & 3.94  & 3.62  & 3.73  & 4.60  & 3.97  & 4.01  & 3.99 \\
    Our RCMNet &       & \textbf{3.18} & \textbf{344.40} & 4.30  & 4.33  & 4.50  & 4.05  & 4.06  & 4.27  & 4.15  & 4.12  & 3.34  & 3.89  & 4.59  & 3.77  & 3.93  & 4.11 \\
    \bottomrule
    \end{tabular}%
    }
  \label{tab:result1}%
\end{table*}%



We introduce the ETSM (\textbf{E}SD \textbf{T}rajectory and confidence map-based \textbf{S}afety \textbf{M}argin) dataset, a specialized collection designed for endoscopic submucosal dissection (ESD) research. This dataset includes annotations for both dissection trajectories and safety margins. Using our custom robotic system~\cite{gao2024transendoscopic}, we captured 21 videos of complete robotic ESD procedures on ex-vivo porcine models. The recordings were made with a flexible dual-channel endoscope (Smart GS-60DQ, HUACO, China) at 30 FPS and a resolution of $1920 \times 1080$. After cropping, the final resolution of the endoscopic images is $ 1300 \times 1024$.

To evaluate our approach, we selected $1,849$ short clips that required dissection trajectory suggestions from $21$ videos. These clips were then divided into two sets. Video $050744$, Video $030204$, Video $000032$, and Video $103115$ are split as the test set, including $369$ short clips. The rest of the videos are used as the training set, including $1480$ short clips. Expert endoscopists from Qilu Hospital annotate the dissection trajectory and safety margin for the last frame of each clip in the annotation software named LabelBox\footnote{\url{https://labelbox.com/}}. Fig.~\ref{fig:dataset} demonstrates our ETSM frames. These frames are annotated with the optimal dissection trajectory, indicated by the red curve, and the safety margin, represented by a blue area. The safety margin, depicted with a gradient effect, is generated by our proposed confidence distributions approach. 



\begin{table}[t]
  \centering
  \caption{Quantitative results of dissection trajectory suggestion.}
    \begin{tabular}{c|ccc}
    \toprule
    Model & ADE $\downarrow$   &  FDE $\downarrow$ & FD $\downarrow$ \\
    \hline
    iDiff-IL & 11.2305  & 13.1144  & 23.4284 \\
    MID   & 9.8655  & 12.3849  & 22.3358 \\
    BC    & 9.1048  & 11.2015  & 20.1367 \\
    \bottomrule
    \end{tabular}%
  \label{tab:result2}%
\end{table}%

\begin{figure*}[h]
    \centering
    \includegraphics[width=0.7\linewidth]{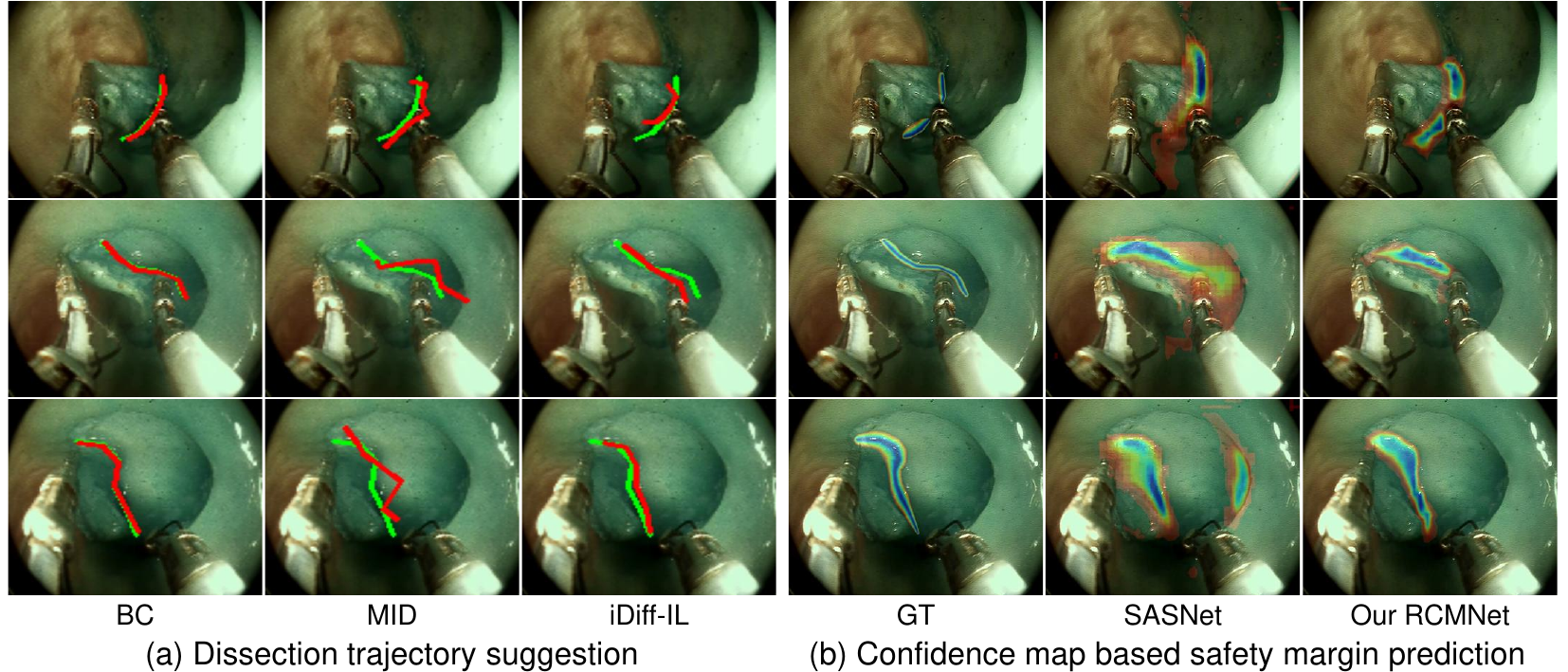}
    \caption{Results visualization. (a) Dissection trajectory suggestion results. Green lines represent the ground truth dissection trajectories, while red lines indicate the predicted trajectories. (b) Confidence map-based safety margin prediction results. }
    \label{fig:result_visu}
\end{figure*}

\subsection{Implementation Details}
We implement our models using the PyTorch framework and conduct all experiments on a single NVIDIA RTX 4090 GPU. 
\textbf{Dissection Trajectory Suggestion:}
We resample the series of 2D coordinate points representing the ground truth trajectory to $6$ points, ensuring that this process does not introduce any unexpected changes in the dissection trajectory trend. The batch size is set to $32$ for images of resolution $128 \times 128$. We train all models for $200$ epochs using the learning rate of $ 0.0001$. The video frames input to the model consists of $3$ frames, representing $3$ seconds.
\textbf{Confidence Map-Based Safety Margin Prediction:} The images and their corresponding confidence maps are resized to either 224 $\times$ 224 or 532 $\times$ 532. A pretrained DINOv2~\cite{oquab2023dinov2} checkpoint is loaded for the encoder, while the decoder is randomly initialized. The DINOv2 encoder is fine-tuned using LoRA~\cite{hu2022lora}, and  the decoder is trained from scratch with a learning rate of 0.001. The batch size is set to $8$ for images of resolution $224 \times 224$ and $4$ for $532 \times 532$. We train the models for 100 epochs using the Adam~\cite{KingBa15} optimizer with a Cosine Annealing scheduler. 

\subsection{Experimental Results and Analysis}
\subsubsection{Dissection Trajectory Suggestion}
To assess the performance of our proposed method, we utilize several evaluation metrics commonly employed in trajectory prediction, as referenced in~\cite{mohamed2020social,sun2020recursive,gu2022stochastic,li2023imitation}. These include the Average Displacement Error (ADE), which measures the overall deviation between predicted and actual trajectories, and the Final Displacement Error (FDE), which calculates the L2 distance between the final points of the predicted and actual trajectories. Additionally, we use the Fréchet Distance (FD) to gauge the geometric similarity between two temporal sequences. All metrics are computed in pixel units, with the input images having a resolution of $128 \times 128$.

Table~\ref{tab:result2} shows that the BC method achieves significantly lower values across all three metrics, indicating a high level of trajectory prediction accuracy. This enhanced precision can greatly contribute to improving the overall outcome and safety of surgical procedures. Fig.~\ref{fig:result_visu}(a) demonstrates the prediction results from the dissection trajectory suggestion module. The predicted trajectory, represented by the red lines, aligns well with the ground truth trajectory, depicted by the green lines.

\subsubsection{Confidence Map-Based Safety Margin Prediction}




To evaluate the performance of our models, we adopt two typical evaluation metrics of regression task, Mean Absolute Error (MAE) and Mean Square Error (MSE), as referenced in~\cite{zhang2015cross, chicco2021coefficient, meng2021cornet, lin2024multidimensional}. Our RCMNet-v1 utilizes the simple convolution layer-based decoder, while our RCMNet-v2 employs the Fully Convolutional Network (FCN) based decoder. Our RCMNet utilizes an All-MLP-based decoder. We compared our RCMNet family of models with
MAN~\cite{lin2022boosting} and SASNet~\cite{sasnet}. As shown in Table~\ref{tab:result1}, our method outperforms these baselines on the test set, achieving lower MAE and MSE at both 224×224 and 532×532 input resolutions. Furthermore, Our RCMNet family of models consistently yield lower MAE under various corruption scenarios, delivering the lowest mean MAE at 224×224 and 532×532, which highlights the robustness of our models. Fig.~\ref{fig:result_visu}(b) demonstrates the prediction results from the confidence map-based safety margin prediction module. The color gradient in the predicted confidence map closely resembles that of the ground truth confidence map. Even in cases where the two dissection trajectories (see the right side of the first row in Fig.~\ref{fig:result_visu}(b)) are more challenging, the prediction results remain satisfactory.
 

\begin{figure}[t]
    \centering
    \includegraphics[width=1\linewidth, trim=0 50 0 0]{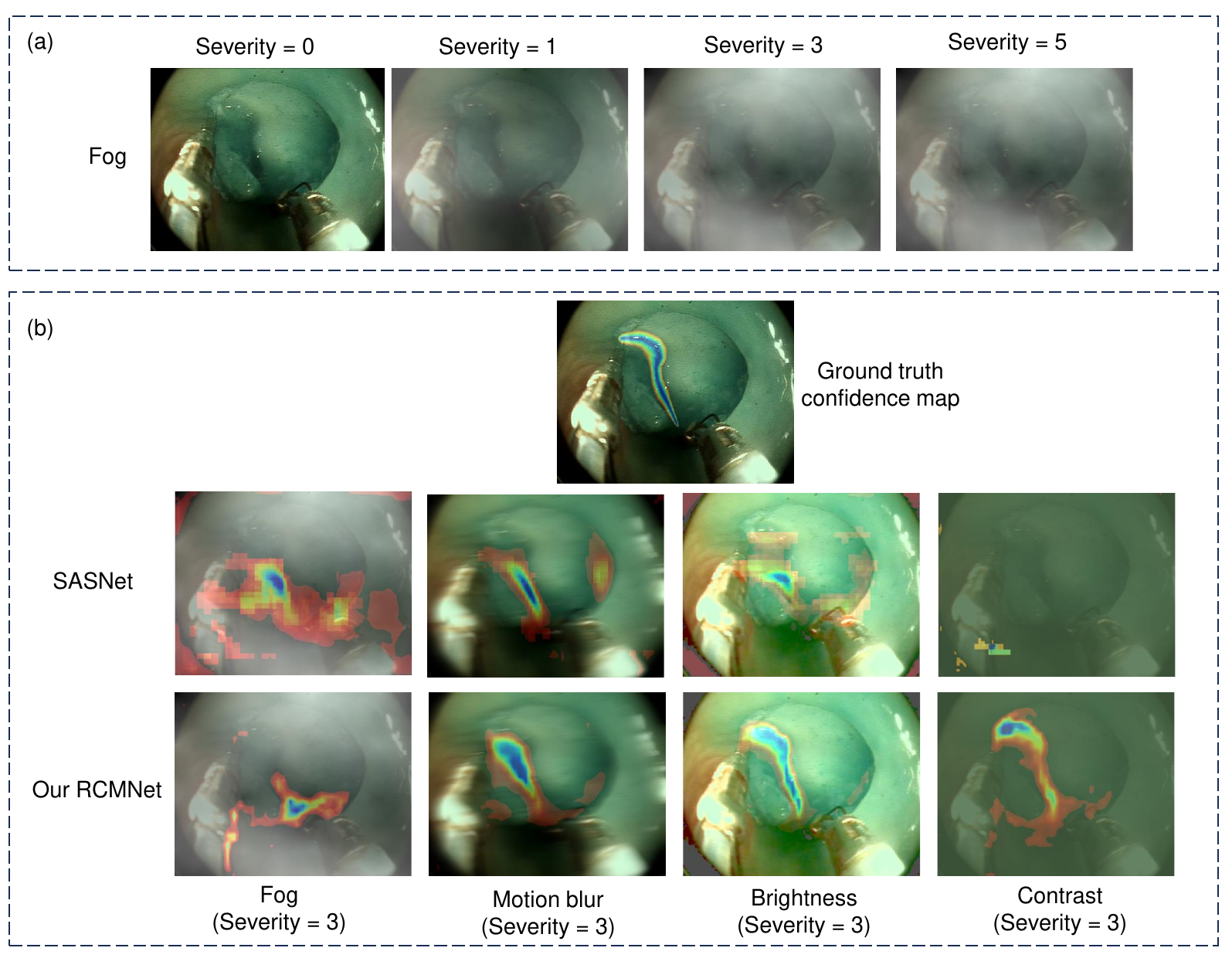}
    \caption{Robustness evaluation under typical data corruptions.}
    \label{fig:robustness}
\vspace{-0.6cm}
\end{figure}

\begin{figure}[t]
\centering
\includegraphics[width=1\linewidth, trim=0 50 0 0]{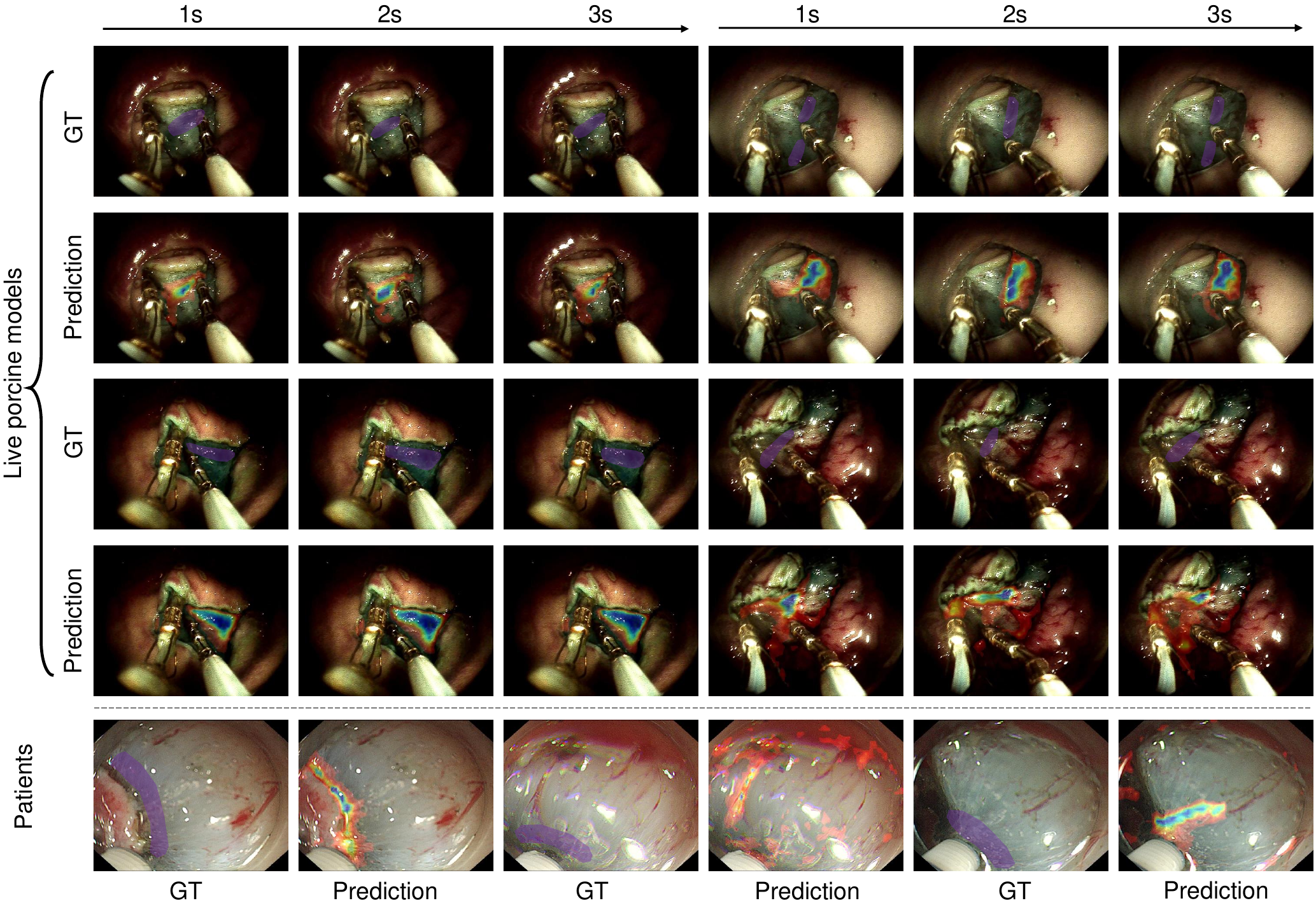}
\caption{Out of domain evaluation. We assess our RCMNet on two additional domains: live porcine data and patient data from Qilu Hospital.}
\label{fig:out_of_domain}
\vspace{-0.4cm}
\end{figure}

\subsubsection{Robustness Evaluation} 
We developed four categories of image corruption (Noise, Blur, Weather, and Digital) with corruption severity ranging from 1 to 5 to evaluate the robustness of methods. A model that maintains higher accuracy despite escalating levels of corruption is deemed more robust~\cite{hendrycks2019benchmarking}. Fig.~\ref{fig:robustness}(a) shows the corrupted image after applying the weather (fog) type. This mimics the smoke produced after cauterizing the tissue. Fig.~\ref{fig:robustness}(b) demonstrates the predicted confidence map on the corrupted images.


\subsubsection{Out-of-domain Evaluation} We conducted a qualitative assessment of out-of-domain data,as shown in Fig.~\ref{fig:out_of_domain}. Compared to the ground truth dissection safety margins (indicated in purple), the confidence maps predicted by our RCMNet are relatively reliable, demonstrating strong generalization.

\subsubsection{Segmentation} We also investigated predicting safety margins as a segmentation task. However, state-of-the-art segmentation methods ~\cite{strudel2021, hong2021deep, SETR} achieved a low IoU score of around 0.45, making the predicted safety margin unreliable.


\section{Conclusion}
In this work, we present advancements in surgical guidance by developing a framework that integrates dissection trajectory suggestions with confidence map-based safety margin predictions, offering vital visual support for surgeons and enabling potential automation of dissection tasks. We also create the ETSM dataset, derived from ESD videos captured by the DREAMS system, which includes detailed annotations for both dissection trajectories and safety margins. Our angular-difference-based algorithm generates confidence maps that guide surgeons by visualizing safety margins effectively. Additionally, we proposed the RCMNet, a novel network architecture combining a Transformer-based image encoder with an All-MLP regression decoder, enhancing prediction accuracy. Collectively, these contributions lay a foundation for improved surgical precision and automation. \textbf{Future work} will focus on extending the RCMNet to incorporate temporal information, enabling the model to leverage sequential data and further improve performance. 
\balance
\bibliographystyle{IEEEtran}
\bibliography{mybib}

\end{document}